\documentclass[11pt]{article}

\usepackage{acl2015}
\usepackage{times}
\usepackage{url}
\usepackage{latexsym}

\usepackage[utf8]{inputenc}
\usepackage[T1]{fontenc}
\usepackage{lmodern}
\usepackage{microtype}
\usepackage{placeins} 
\usepackage{amsmath}
\usepackage{graphicx}
\usepackage{xcolor}
\usepackage[most]{tcolorbox}
\usepackage[numbers]{natbib}

\usepackage{tikz}
\usetikzlibrary{
    arrows.meta,
    shapes.geometric,
    shapes,
    arrows,
    positioning,
    fit,
    calc
}

\usepackage{fontawesome5}

\title{Redefining CX with Agentic AI: Minerva CQ Case Study}

\author{
\begin{tabular}{c}
Garima Agrawal \quad Riccardo De Maria \quad Kiran Davuluri \quad Daniele Spera \\
Charlie Read \quad Cosimo Spera \quad Jack Garrett \quad Don Miller\thanks{All authors contributed equally to this work. This work was conducted as part of the Minerva CQ AI Research Division.} \\
Minerva CQ, California, USA
\end{tabular}
\\
\texttt{\{garima, riccardo.demaria, kdavuluri, daniele\}@minervacq.com} \\
\texttt{\{cread, cosimo, jack, dmiller\}@minervacq.com}
}

\date{}

\begin{document}
\maketitle

\begin{abstract}
Despite advances in AI for contact centers, customer experience (CX) continues to suffer from high average handling time (AHT), low first-call resolution, and poor customer satisfaction (CSAT). A key driver is the cognitive load on agents, who must navigate fragmented systems, troubleshoot manually, and frequently place customers on hold. Existing AI-powered agent-assist tools are often reactive—driven by static rules, simple prompting, or retrieval-augmented generation (RAG) without deeper contextual reasoning.

We introduce \textbf{Agentic AI}: goal-driven, autonomous, tool-using systems that proactively support agents in real time. Unlike conventional approaches, Agentic AI identifies customer intent, triggers modular workflows, maintains evolving context, and adapts dynamically to conversation state.

This paper presents a case study of \textbf{Minerva CQ}, a real-time Agent Assist product deployed in voice-based customer support. Minerva CQ integrates real-time transcription, intent and sentiment detection, entity recognition, contextual retrieval, dynamic customer profiling, and partial conversational summaries—enabling proactive workflows and continuous context-building. Deployed in live production, Minerva CQ acts as an AI co-pilot, delivering measurable improvements in agent efficiency and customer experience across multiple deployments.
\end{abstract}

\section{Introduction}

Customer experience (CX) has become a defining differentiator across industries—from telecommunications and retail to healthcare and automotive~\cite{arkadan2024cxo}. Within this landscape, \textbf{contact centers and service teams} serve as the frontline of brand engagement. Yet despite increasing investments in automation and AI, service quality often remains inconsistent. Persistent challenges include:

\begin{itemize}
    \item High average handling time (AHT) due to manual troubleshooting and fragmented tool usage
    \item Customer frustration from being placed on hold or repeating information
    \item Agent overload navigating complex workflows and unstructured knowledge bases (KBs)
    \item Variability in AI-generated answers due to lack of structure and contextual awareness in LLM/RAG systems
    \item Low first-call resolution (FCR) and declining customer satisfaction (CSAT) and Net Promoter Score (NPS)
\end{itemize}

To address these challenges, many enterprises have adopted \textbf{agent-assist systems} designed to surface relevant answers, summarize interactions, or suggest next steps~\cite{ye2025sopagent, yang2025contextagent,sapkota2025ai}. However, most existing solutions are inherently \textit{reactive}—they respond only when prompted, lack temporal awareness, and fail to support agents proactively throughout the customer journey.

\subsection*{Limitations of Reactive Agent-Assist Systems}

Today’s agent-assist tools typically fall into three categories:

\begin{itemize}
    \item Rule-based or decision-tree systems that are rigid and prone to failure under conversational drift
    \item Prompt-based LLMs that generate plausible responses but lack continuity and memory
    \item Retrieval-Augmented Generation (RAG) systems that fetch documents from a KB based on queries, but rely on explicit prompting and offer limited contextual integration
\end{itemize}

These solutions are passive by design. They do not track user goals, update internal memory, or orchestrate multi-step reasoning. As a result, the cognitive burden remains on the human agent to interpret outputs and determine next actions—especially under real-time pressure.

\subsection*{The Shift Toward Agentic AI}

To move beyond the limitations of reactive systems, we advocate for a new paradigm: \textbf{Agentic AI}—intelligent, autonomous, goal-driven systems that operate with continuity and intent. Agentic AI systems do not merely respond; they \textit{act}. They infer intent, maintain evolving memory, plan actions, and invoke tools in service of customer outcomes. While the term “agentic” is gaining popularity, true agentic behavior requires integrated planning, workflow orchestration, and sustained contextual reasoning—not just LLMs augmented with tools.

\subsection*{Minerva CQ: Real-Time Agent-Assist, Not Just RAG}

We present a case study of \textbf{Minerva CQ}, a real-time agent-assist product purpose-built for voice-based customer support. Unlike standalone RAG platforms, Minerva CQ leverages RAG selectively as one component within a broader agentic framework. The system integrates multilingual automatic speech recognition (ASR), dynamic intent and sentiment detection, autonomous workflow triggering, entity tracking, contextual profiling, and live conversation state management.

Minerva CQ incorporates several distinctive capabilities that elevate it beyond traditional agent-assist tools. The system automatically identifies customer intent from live conversations and generates AI-suggested queries, allowing agents to retrieve answers without manual typing or search. Frequently asked questions—mined from both historical transcripts and live queries—are validated and cached to provide instant responses when applicable, reducing latency, cost, and reliance on LLMs. It also generates compact, incremental \textbf{partial summaries} that preserve and evolve the context across conversation turns, supporting downstream reasoning and recommendations. Additionally, Minerva CQ surfaces intent-aware workflows and next-best actions to guide agents in real time, and concludes each interaction with an automatic call summary, CSAT/NPS estimation, and sentiment tracking throughout the call.

By embedding Agentic AI principles across the full interaction pipeline, Minerva CQ transforms agent assist into a proactive AI co-pilot—enhancing both agent efficiency and customer satisfaction in high-stakes, real-time environments. Notably, Minerva CQ has demonstrated success in both \textbf{customer service and sales scenarios}, dynamically adapting to the tone, intent, and goals of the interaction. This flexibility enables consistent support across a wide range of workflows, from resolving technical issues to driving conversions.

\subsection*{Contributions}
The key contributions of this paper are:
\begin{itemize}
    \item We present a deployed agentic AI system for real-time voice-based contact centers, integrating multilingual ASR, contextual understanding, and workflow automation.
    \item We describe a novel combination of partial conversational summaries, proactive query generation, validated FAQ caching, and intent-triggered workflows that reduce cognitive load and operational latency.
    \item We report results from live production pilots demonstrating measurable improvements in AHT, FCR, conversion rates, and customer satisfaction metrics.
\end{itemize}

The remainder of this paper outlines our system design, operational insights, and the practical implications of deploying Agentic AI at scale in enterprise contact centers.

\section{End-to-End Agentic AI Workflow for Customer Experience}
\label{sec:workflow}

Minerva CQ’s Agentic AI platform is designed to transform customer experience (CX) by supporting human agents throughout the full lifecycle of a conversation—proactively, intelligently, and in real time. Unlike traditional systems that offer isolated suggestions or reactive responses, Minerva CQ functions as an integrated, goal-driven co-pilot, embedded seamlessly within live voice-based interactions.

This section outlines the key phases of a typical customer call, showing how Minerva CQ’s agentic capabilities align with natural touchpoints in the agent–customer journey. Each system component is described along with its motivation and design rationale.

\subsection{Call Initiation: Establishing Customer Context}

At call start, the system identifies the caller and retrieves relevant CRM data. Beyond static context fetching, Minerva CQ’s entity recognition module parses the live transcript to extract identifiers (e.g., name, email, account number) as they are mentioned. This eliminates repetitive verification questions, accelerates authentication, and improves early engagement.

\subsection{Intent Recognition and Workflow Triggering}

As the conversation progresses, Minerva CQ continuously analyzes the transcript to detect underlying customer intent. Once intent confidence is high, it proactively triggers the corresponding workflow—such as a billing correction or plan change—and surfaces next-best actions tailored to that scenario. This removes the need for manual search, enabling faster and more consistent resolution.

\subsection{Customer Profiling in Sales Conversations}

For outbound or sales-oriented calls, Minerva CQ passively builds a behavioral profile by detecting expressions of interest, hesitation, or goal-oriented language. These cues feed into lightweight dynamic profiling, enabling tailored offer recommendations, upsell paths, or probing questions—supporting conversions without breaking conversational flow.

\subsection{Context-Aware Knowledge Querying: Beyond Keyword-Based RAG}

Navigating fragmented knowledge bases (KBs) mid-conversation is a major agent pain point. Traditional keyword lookups via portals, PDFs, or spreadsheets slow down calls and increase cognitive load.  

While many modern agent-assist tools integrate RAG search bars, they still rely on agents to manually craft queries—an unrealistic expectation under real-time pressure. Poorly formed queries further degrade retrieval quality.

Minerva CQ automates this step by generating semantically rich, context-grounded queries directly from the live transcript. It identifies implicit or explicit customer asks and formulates well-structured, KB-compatible questions, presented as clickable prompts.  

To reduce latency and cost, Minerva CQ maintains a validated FAQ cache mined from transcripts and live usage. Matching queries retrieve instant responses, bypassing RAG entirely. New FAQs undergo heuristic and ontology checks before being added to the cache.

\begin{tcolorbox}[colback=gray!10, colframe=gray!40, boxrule=0.5pt, arc=2pt, left=4pt, right=4pt, top=4pt, bottom=4pt]
\textbf{Example:} If a customer says, \textbf{``I want to get a travel plan''}, a naive system might ask \textbf{``Which offer is the customer asking about?''}—too vague for KB retrieval. Minerva CQ reformulates this into actionable queries such as \textbf{``Which travel offers are available?''} or \textbf{``How to activate a travel plan?''}, bridging conversational language and structured KB content.
\end{tcolorbox}

\subsection{Partial Summarization: Maintaining Conversation Context}

Minerva CQ produces compact, incremental \textbf{partial summaries} throughout the call. These evolve turn-by-turn, capturing salient facts while filtering noise—helpful for both agents (quick reference) and downstream modules (maintaining cross-turn context for reasoning, workflow triggers, and sentiment analysis).

\subsection{Agent Guidance: Sentiment, Intent, and Interaction Indicators}

In parallel, Minerva CQ tracks sentiment shifts, intent changes, and CSAT/NPS likelihood in real time. Indicators appear on the live agent dashboard, enabling on-the-fly adjustments in tone, pacing, and phrasing—critical for de-escalation, upselling, and emotionally sensitive situations.

\subsection{Call Conclusion and Compliance}

At call end, Minerva CQ auto-generates a final summary including primary intent, resolution path, agent actions, and sentiment trajectory. Personally identifiable information (PII) is redacted to ensure compliance and reduce post-call documentation.

\subsection{A Unified Agentic Framework}

Minerva CQ orchestrates these capabilities through an \textit{observe} → \textit{understand} → \textit{decide} → \textit{act} → \textit{assist} → \textit{learn} loop. Unlike modular tools that operate in isolation, Minerva CQ connects each stage to sustain momentum toward resolution. Its proactive, goal-oriented design reduces agent burden while enhancing consistency and personalization.

\paragraph{Key Features Summary.} 
The following capabilities collectively distinguish Minerva CQ as a real-time agent-assist solution built on Agentic AI principles. Their operational benefits and KPI impact are summarized in Table~\ref{tab:feature-impact} in Section~\ref{sec:roi}:

\begin{itemize}
    \item Real-time entity extraction, intent recognition, and dynamic workflow triggering
    \item Proactive AI-suggested query generation with validated FAQ caching
    \item Partial conversation summaries for context retention
    \item Live sentiment/CSAT/NPS tracking and automated final summaries
\end{itemize}

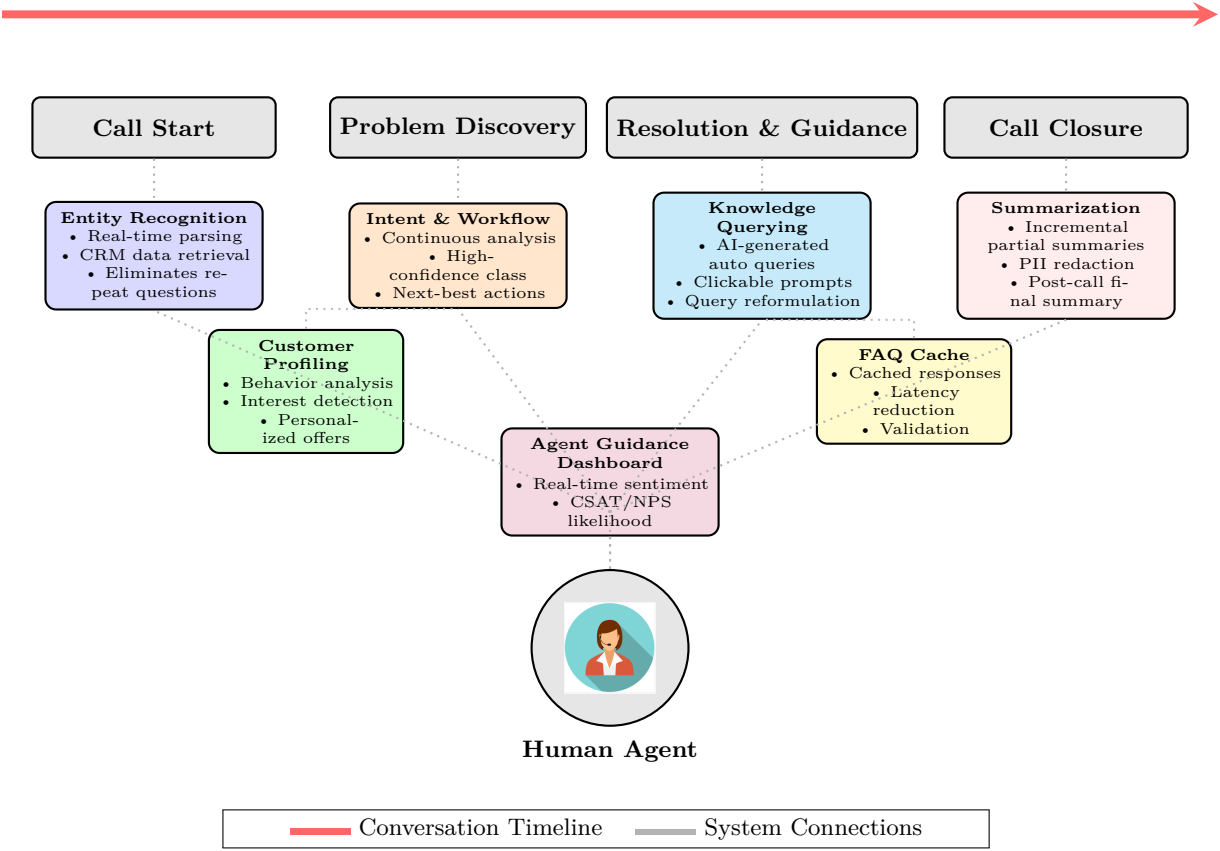
\begin{figure*}[t]
\centering
\begin{tikzpicture}[
    phase/.style={rectangle, draw=black, thick, fill=gray!20, minimum width=3.2cm, minimum height=0.8cm, 
                 rounded corners=3pt, font=\bfseries\footnotesize, align=center},
    module/.style={rectangle, draw=black, thick, minimum width=2.8cm, minimum height=1.2cm, 
                  rounded corners=4pt, align=center, font=\tiny, text width=2.6cm},
    support_module/.style={rectangle, draw=black, thick, minimum width=2.5cm, minimum height=1.0cm, 
                          rounded corners=4pt, align=center, font=\tiny, text width=2.3cm},
    timeline_arrow/.style={->, very thick, >=stealth, color=red!60},
    connection/.style={thick, color=gray!60, dotted},
    sub_connection/.style={thick, color=gray!50, dotted}
]

\draw[timeline_arrow, line width=3pt] (0,8) -- (16,8);
\node[above] at (8,8.2) {\textbf{Conversation Timeline}};

\node[phase] (p1) at (2,6.5) {Call Start};
\node[phase] (p2) at (6,6.5) {Problem Discovery};
\node[phase] (p3) at (10,6.5) {Resolution \& Guidance};
\node[phase] (p4) at (14,6.5) {Call Closure};

\node[module, fill=blue!15] (m1) at (2,4.8) {
    \textbf{Entity Recognition}\\
    • Real-time parsing\\
    • CRM data retrieval\\
    • Eliminates repeat questions
};

\node[module, fill=orange!20] (m2) at (6,4.8) {
    \textbf{Intent \& Workflow}\\
    • Continuous analysis\\
    • High-confidence class\\
    • Next-best actions
};

\node[module, fill=cyan!20] (m3) at (10,4.8) {
    \textbf{Knowledge Querying}\\
    • AI-generated auto queries\\
    • Clickable prompts\\
    • Query reformulation
};

\node[module, fill=pink!30] (m4) at (14,4.8) {
    \textbf{Summarization}\\
    • Incremental partial summaries\\
    • PII redaction\\
    • Post-call final summary
};

\node[support_module, fill=green!20] (m5) at (4,3.0) {
    \textbf{Customer Profiling}\\
    • Behavior analysis\\
    • Interest detection\\
    • Personalized offers
};

\node[support_module, fill=yellow!25] (m6) at (12,3.0) {
    \textbf{FAQ Cache}\\
    • Cached responses\\
    • Latency reduction\\
    • Validation
};

\node[module, fill=purple!15] (m7) at (8,1.8) {
    \textbf{Agent Guidance}\\
    \textbf{Dashboard}\\
    \vspace{1pt}
    • Real-time sentiment\\
    • CSAT/NPS likelihood
};

\node[draw, thick, circle, fill=gray!20, minimum size=1.6cm] (agent) at (8,-0.4) {
    \includegraphics[width=1.2cm]{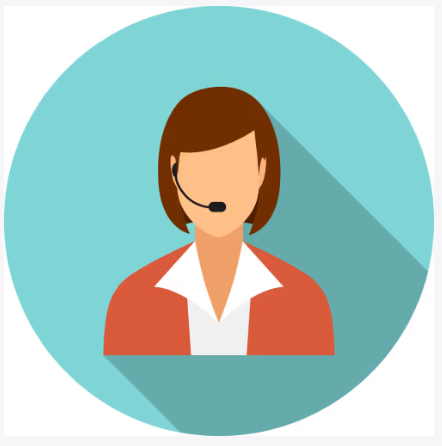}
};
\node[below=0.05cm of agent, font=\small\bfseries] {Human Agent};

\draw[connection] (p1) -- (m1);
\draw[connection] (p2) -- (m2);
\draw[connection] (p3) -- (m3);
\draw[connection] (p4) -- (m4);

\draw[sub_connection] (m5.north) -- (m5.north |- m2.south) -- (m2.south);
\draw[sub_connection] (m6.north) -- (m6.north |- m3.south) -- (m3.south);

\coordinate (hub) at (8,1.4);
\draw[connection] (agent.north) -- (hub);

\draw[connection] (hub) -- (m1.south);
\draw[connection] (hub) -- (m2.south);
\draw[connection] (hub) -- (m3.south);
\draw[connection] (hub) -- (m4.south);

\draw[connection] (m7.south) -- (agent.north);

\end{tikzpicture}

\vspace{0.2cm}
\begin{center}
\fbox{
\begin{minipage}{0.6\textwidth}
\centering
\footnotesize
\textcolor{red!60}{\rule{0.8cm}{2.5pt}} Conversation Timeline \quad
\textcolor{gray!60}{\rule{0.8cm}{2pt}} System Connections
\end{minipage}
}
\end{center}

\caption{Minerva CQ's Agentic AI Framework: Human agents are supported by intelligent AI modules that activate during specific conversation phases. The system provides real-time transcript analysis, proactive knowledge retrieval, behavioral profiling, and continuous context preservation—transforming the agent experience from reactive assistance to goal-driven co-piloting.}
\label{fig:minerva-cq-framework}
\end{figure*}

\section{Measuring Feature Impact and ROI}
\label{sec:roi}

While Section~\ref{sec:workflow} detailed Minerva CQ’s multi-phase agentic workflow, this section examines how specific features translate into measurable operational benefits and business outcomes. Drawing on real-world pilot deployments in a sales usecase, we quantify improvements in agent productivity, customer experience, and revenue metrics.

\subsection{Operational Impact of Key Features}

Table~\ref{tab:feature-impact} maps Minerva CQ’s major agentic features to their operational benefits and the KPIs they most directly influence. This condensed view connects the system capabilities described in Section~\ref{sec:workflow} to downstream performance measures without repeating the detailed descriptions.

\begin{table*}[t]
\centering
\small
\begin{tabular}{p{3.5cm} p{7cm} p{4.5cm}}
\hline
\textbf{Feature} & \textbf{Operational Benefit} & \textbf{KPI Impact} \\
\hline
Entity Recognition \& CRM Context Retrieval & 
Reduces repetitive verification questions and accelerates call initiation & 
Lower AHT \\

Intent Recognition \& Workflow Triggering & 
Guides agents with next-best actions; enables proactive resolution & 
Lower AHT, Higher FCR \\

AI Query Generation \& FAQ Cache & 
Eliminates manual search; provides validated, low-latency responses & 
Lower AHT, Consistent CX \\

Partial \& Final Summarization & 
Reduces after-call work; supports compliance with PII redaction & 
Lower Wrap-up Time, Improved Compliance \\
\hline
\end{tabular}
\caption{Mapping Minerva CQ features to operational benefits and corresponding KPI improvements.}
\label{tab:feature-impact}
\end{table*}

\subsection{Pilot ROI Analysis}

\paragraph{Evaluation Methodology.}  
The pilot was structured as a live production A/B test with two cohorts of 50 agents each: one group using Minerva CQ’s agentic AI, the other operating without AI support. Both handled the same mix of inbound service and sales calls under identical routing rules and product offerings. The evaluation covered approximately 40{,}000 production voice interactions over the pilot period. KPIs—Average Handling Time (AHT), Lead-to-Enquiry (L2E) conversion, and booking conversion (enquiry→booking)—were computed from call logs and CRM outcomes. While seasonality and time-of-day effects were not fully eliminated, the balanced cohort design reduces major confounders.

\paragraph{Key ROI Insights.}  
\begin{itemize}
    \item \textbf{38\% reduction in Average Handling Time (AHT)} – AHT reduced from \textbf{4m 43s to 2m 55s}, enabling faster call resolution and lower cost-to-serve.
    \item \textbf{33\% uplift in Lead-to-Enquiry (L2E) conversion} – Proactive guidance and AI-suggested queries improved pipeline efficiency.
    \item \textbf{4.8\% uplift in booking conversion} – Consistent, validated responses and contextual guidance resulted in measurable revenue impact.
\end{itemize}

\begin{figure*}[t]
  \centering
  \includegraphics[width=0.85\textwidth]{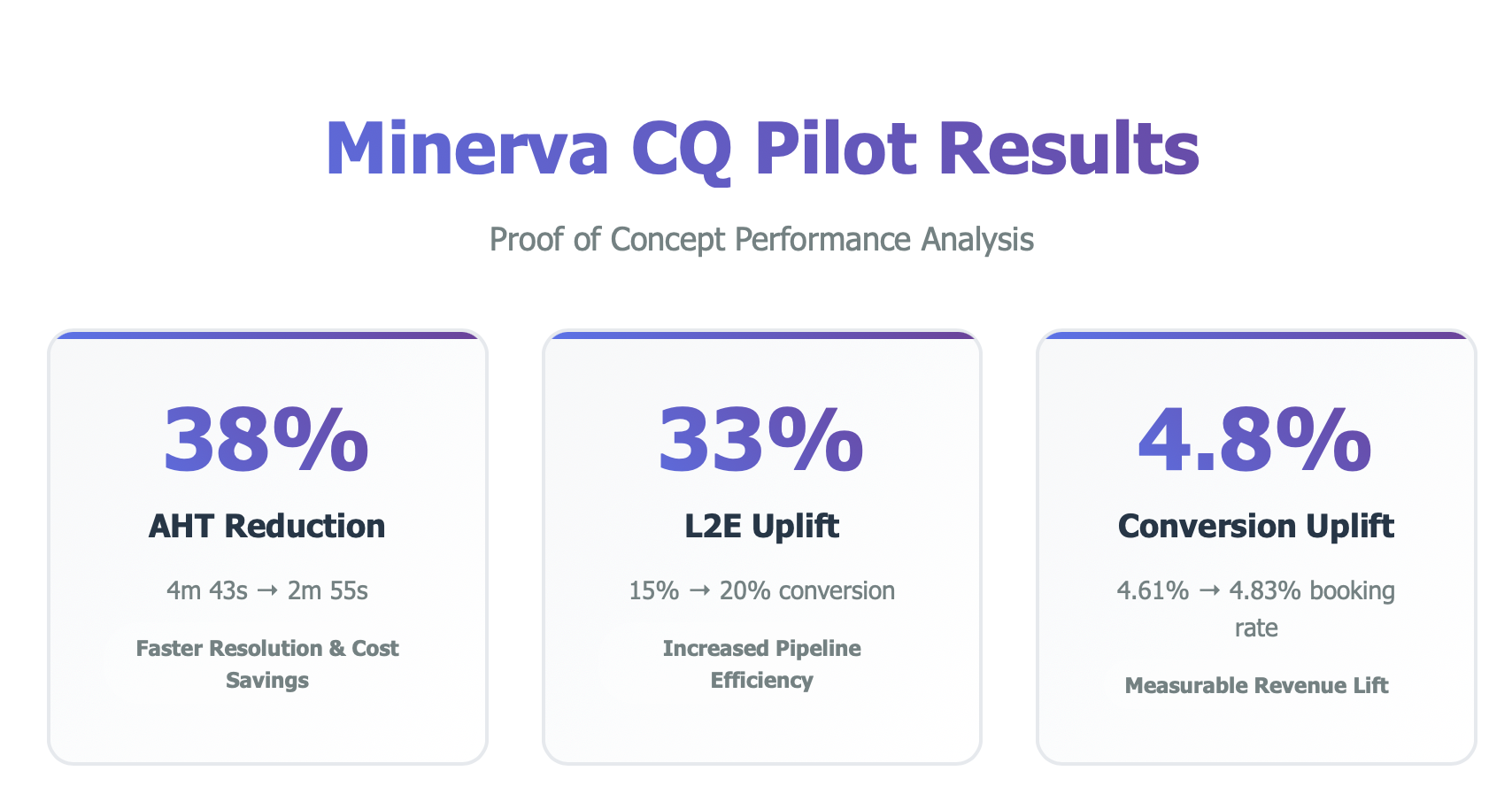}
  \caption{Key performance improvements with Minerva CQ in a live client case study:
  38\% reduction in Average Handling Time (AHT), 33\% uplift in Lead-to-Enquiry (L2E),
  and 4.8\% uplift in booking conversion. Client details withheld; results from
  production voice interactions.}
  \label{fig:minerva-kpi}
\end{figure*}

As shown in Figure~\ref{fig:minerva-kpi}, these improvements are reflected in all three KPIs, with Minerva-assisted agents outperforming the control group across efficiency and conversion measures.

\paragraph{Latency and Cost Savings from FAQ Matching.}  
FAQ matching delivered both speed and cost advantages. In deployment logs from 10{,}000 calls, approximately 7{,}000 queries were answered from the validated FAQ cache instead of invoking RAG. Each avoided RAG call saved agents an average of 6 seconds of wait time (cumulative $\sim$11.7 hours latency saved) and reduced LLM API usage, lowering inference costs. Traditional RAG retrieval via OpenSearch typically took 5–9 seconds, while FAQ matching returned results in under 0.5 seconds—helping to compress handling times and improve customer experience.

\begin{figure*}[t]
  \centering
  \includegraphics[width=0.85\textwidth]{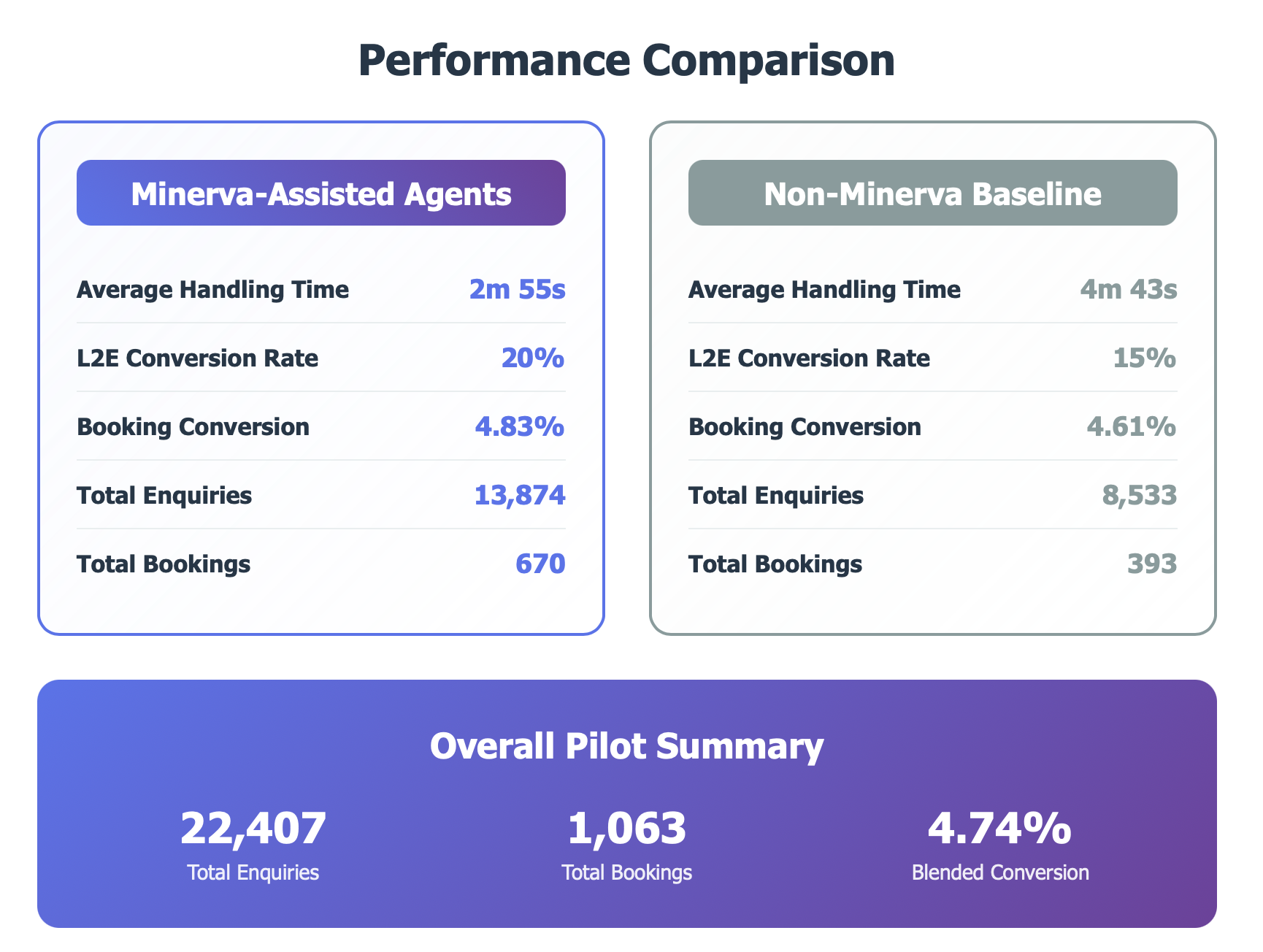}
  \caption{Minerva-assisted vs.\ non-assisted performance in the same environment:
  faster handling time, higher L2E and booking conversion, and stronger overall volume metrics.
  Totals shown are anonymized aggregates from the pilot period.}
  \label{fig:minerva-comparison}
\end{figure*}

\paragraph{Multilingual Capability.}  
Minerva CQ supports multilingual and mixed-language conversations in live deployments, including \textbf{English} for US contact centers, \textbf{Italian} and other European languages for EU contact centers, and \textbf{Hindi/Hinglish} for India in Asia. In the India deployment, the client setup has no language flag for outbound calls—customers may speak Hindi, English, or switch freely between the two. Agents can speak both languages, but not all can read Hindi. To address this, Minerva CQ’s speech-to-text pipeline:
\begin{enumerate}
    \item Accurately transcribes Hindi, Indian English, or Hinglish speech in real time.
    \item Converts transcriptions into English text for AI-suggested prompts, ensuring all agents can read suggestions without delay.
\end{enumerate}
This design enables true code-switching support without introducing additional latency, maintaining real-time AI guidance in dynamic multilingual settings.

\paragraph{Summary.}  
Integrating Minerva CQ into live voice workflows led to:
\begin{itemize}
    \item Reduced operational cost through faster call handling
    \item Improved pipeline engagement and lead conversion
    \item Measurable uplift in revenue-driving KPIs
\end{itemize}
These results validate the transition from \textbf{reactive agent-assist} to \textbf{proactive agentic AI co-piloting}.

\section{Related Work}

\noindent \textbf{Customer experience and AI-assisted contact centers -}
Field studies show that AI assistance can improve service productivity at scale. For example, a randomized deployment of a generative assistant to over 5{,}000 agents increased issues resolved per hour by 14--15\% \citep{brynjolfsson2025generative}. Contact-center research also documents persistent agent challenges such as manual knowledge lookups, after-call documentation, and fragmented tool usage, motivating real-time transcription and summarization to reduce cognitive load \citep{sachdeva2023realtime}. While adoption is growing, most deployed tools remain \emph{reactive}: they respond only when prompted and lack sustained conversation-level state or proactive guidance.

\noindent \textbf{From reactive to proactive, context-aware assistance -}
A key capability in proactive systems is conversational query reformulation: under-specified utterances must be rewritten into self-contained queries for accurate retrieval. Manual rewrites in TREC CAsT significantly outperform automatic baselines, highlighting the importance of robust reformulation \citep{elgohary2019canard,dalton2020cast}. 

\noindent \textbf{Retrieval-augmented generation and its trade-offs -}
Retrieval-Augmented Generation (RAG) can improve grounding but is highly sensitive to retrieval quality and can introduce latency and cost in production settings \citep{salemi2024erag}. This makes frequent or repeated questions an ideal target for low-latency alternatives.

\noindent \textbf{FAQ retrieval for efficiency -}
FAQ retrieval systems offer such an alternative, returning curated, validated answers with millisecond-level latency. Early methods relied on query clustering \citep{sakata2019faq}, while recent work uses dense bi-encoder models with intent-aware matching \citep{chen2023intentfaq}. 

Our system builds on these strands by (i) proactively generating KB-compatible queries from the live transcript, (ii) serving validated FAQ answers to reduce RAG calls and cost \citep{agrawal2024beyondRAG}, and (iii) maintaining partial, streaming summaries to preserve conversational context across turns .

\section{Discussion}

\subsection{From Reactive Assistance to Agentic Co-Piloting}
This case study demonstrates that shifting from reactive, prompt-driven agent assist to an agentic, goal-oriented architecture measurably improves workflows. Unlike systems that leave interpretation and next-step selection to the agent, Minerva CQ maintains evolving context, plans actions, and invokes tools to advance customer outcomes—reducing cognitive load and variability in real time. Its design integrates entity extraction and CRM retrieval at call start, continuous intent detection with workflow triggering during the call, and automated summarization with compliance at close, ensuring continuity across the interaction lifecycle.

\subsection{Design Drivers Behind KPI Gains}
Two mechanisms stand out as primary contributors to the observed improvements:
\begin{enumerate}
    \item \textbf{Proactive, KB-compatible query generation with validated FAQ caching:} Reformulating under-specified customer asks into retrievable queries removes agent query-writing effort and mitigates a known RAG failure mode. The validated FAQ cache returns less than 0.5\,s responses, avoiding costly RAG calls. In 10k calls, $\sim$7k queries were served from cache, saving $\sim$11.7 hours of cumulative latency.
    \item \textbf{Partial, streaming summaries with workflow guidance:} Turn-by-turn summaries preserve state for downstream reasoning, while intent-aware next-best actions reduce decision latency and support consistent execution.
\end{enumerate}
Together, these mechanisms align with the system-to-KPI mapping in Table~\ref{tab:feature-impact} and match the efficiency and conversion gains observed in the A/B pilot.

\subsection{Business Impact in Live Production}
In an A/B pilot covering $\sim$40{,}000 production voice interactions across matched cohorts (50 agents per arm), Minerva-assisted agents achieved a 38\% reduction in AHT (4m43s $\rightarrow$ 2m55s), a 33\% uplift in L2E conversion, and a 4.8\% uplift in bookings. 

Shorter AHT directly correlates with increased agent productivity. When agents spend less time handling each call, they can resolve more customer issues within the same shift duration. This allows the same workforce to handle a higher volume of interactions without increasing headcount—offering a win-win scenario for both operational efficiency and business scalability.

Beyond real-time assistance, Minerva CQ functions as a strategic intelligence layer through its proactive ``voice of the customer'' reporting. By aggregating and analyzing customer pain points, agent challenges, and sentiment trends, these reports provide stakeholders with a holistic view of what is working well and where targeted changes in products, policies, or service flows could yield greater CX gains. This positions Minerva CQ not only as an efficiency driver but also as a business insight engine, enabling organizations to act on evidence-based priorities that extend beyond the call.

These results substantiate the central claim: moving from reactive assistance to agentic co-piloting delivers efficiency, conversion, and intelligence gains in live, voice-based CX environments.

\subsection{Operational Breadth: Multilingual, Code-Switching, Compliance}
The deployment also demonstrates robustness to multilingual, code-switching environments (e.g., Hindi/English/Hinglish), where the ASR pipeline produces readable English prompts for all agents without added latency—maintaining guidance quality despite language shifts. Automated, PII-aware final summaries reduce after-call work while supporting compliance requirements.

\subsection{Positioning Within the Literature and Market}
Minerva’s architecture directly addresses well-documented pain points in contact centers—manual KB lookups, fragmented tools, and after-call documentation—by maintaining conversation-level state and proactively guiding the agent, in line with the broader industry shift from reactive tools to context-aware, workflow-integrated assistance.

The system also occupies a pragmatic point in the RAG design space: retrieval is used selectively, while frequent questions are routed to a validated, low-latency FAQ path. This approach reflects known latency/cost trade-offs of RAG in production and the operational value of purpose-built FAQ retrieval.

\subsection{Limitations}
While the live A/B design improves ecological validity, residual temporal confounders (e.g., seasonality, time-of-day) may influence outcomes. CSAT and NPS remain partly shaped by factors outside Minerva CQ’s direct control, such as agent training, customer history, or offer competitiveness. Minerva mitigates this through ``voice of customer" reporting that surfaces systemic friction points, agent feedback, and sentiment trends.

\subsection{Implications for Industry Track}
For practitioners, the key takeaway is an integrative pattern that blends agentic planning with retrieval pragmatism: (1) proactive query reformulation; (2) FAQ caching as the fast path; (3) partial summaries to maintain state; (4) workflow triggers for next-best actions; and (5) compliance-ready documentation—all embedded in an \emph{observe→understand→decide→act→assist→learn} loop suited to real-time voice. This pattern is both deployable and measurable in ROI terms, while remaining robust to multilingual, code-switching realities of global CX.

\subsection{Future Directions for Industry}
\textbf{Standardized evaluation:} Establish shared, voice-centric evaluation slices (e.g., repeated-FAQ vs.\ novel-policy queries, escalation/de-escalation segments) and report latency distributions alongside accuracy to reflect real-time constraints.

\textbf{FAQ governance and drift:} Extend the validated FAQ lifecycle (mining→validation→cache→expiry) with automatic drift detection and policy/version control, preserving the latency advantage while maintaining correctness.

\textbf{Cross-channel generalization:} Investigate how partial summarization and intent-triggered workflows transfer from voice to chat/email while preserving turnaround-time guarantees and compliance controls.

\textbf{Human factors and training:} Quantify how guidance frequency, timing, and UI presentation affect agent trust, handoff smoothness, and learning curves—particularly in multilingual environments.

\textbf{Cost–latency knobs:} Systematically characterize when to route to FAQ vs.\ RAG vs.\ tool actions under SLA/throughput constraints, extending the selective-retrieval strategy already deployed here.

\paragraph{Summary.} The evidence indicates that an agentic, workflow-integrated approach—anchored by proactive query generation, validated FAQ caching, and persistent conversation state—can deliver both efficiency and revenue benefits at production scale, while also generating strategic intelligence for the business. 
By reducing the average handling time (AHT), the system effectively increases agent throughput—enabling the same team to handle a greater volume of calls without adding headcount. This translates into higher productivity per agent and contributes to scalable growth in support operations.
This positions Minerva CQ as a credible blueprint for real-time, voice-first agent co-pilots in enterprise CX.

In sum, Minerva CQ exemplifies the broader industry shift toward agentic AI in customer experience—moving from reactive assistance to integrated, context-aware co-pilots that not only optimize live interactions but also generate strategic intelligence for continuous business transformation.

\section{Conclusion}
Minerva CQ unifies real-time reasoning, selective retrieval, and workflow automation to support human agents end-to-end in live customer interactions. By maintaining conversational context, proactively generating high-quality queries, leveraging validated FAQ caching, and triggering next-best actions, it addresses persistent contact center challenges of information overload, fragmented tools, and inconsistent query handling.

Deployed at production scale, Minerva CQ delivered measurable efficiency and conversion gains while maintaining robustness in multilingual, code-switching settings and ensuring compliance through automated summaries. By enabling agents to resolve more conversations within the same operational bandwidth, the system boosts individual productivity and supports scalable service delivery. The results highlights a broader industry shift from reactive assistance toward integrated, context-aware co-piloting—where AI systems actively shape customer outcomes in real time and generate strategic intelligence for continuous improvement.

\bibliographystyle{acl_natbib}











\end{document}